\documentclass[review]{elsarticle}

\usepackage{lineno,hyperref}
\modulolinenumbers[5]

\usepackage{subfigure}
\usepackage{lscape}
\usepackage{makecell}
\usepackage{multirow}

\usepackage{amsmath,amssymb} 
\usepackage{graphicx}
\usepackage{algorithm}
\usepackage{algorithmic}
\usepackage{multirow}

\usepackage{bm}
\newcommand{\etal}{\textit{et al.}}

\journal{Neural Network}

\begin{document}

\begin{frontmatter}

\title{Attribute-Specific Manipulation Based on Layer-Wise Channels}

%
%
%

\author[address1,address12]{Yuanjie Yan}
\address[address1]{State Key Laboratory for Novel Software Technology, Nanjing University, China.}
\ead{yanyj@smail.nju.edu.cn}

\address[address12]{Department of Computer Science and Technology, Nanjing University, China.}

\author[address3]{Jian Zhao}
\address[address3]{School of Electronic Science and Engineering, Nanjing University, Nanjing, China.}
\ead{jianzhao@nju.edu.cn}

\author[address1,address2]{Furao Shen\corref{mycorrespondingauthor}}
\cortext[mycorrespondingauthor]{Corresponding author}
\ead{frshen@nju.edu.cn}

\address[address2]{School of Artificial Intelligence, Nanjing University, Nanjing, China.}


\begin{abstract}
Image manipulation on the latent space of the pre-trained StyleGAN can control the semantic attributes of the generated images.
Recently, some studies have focused on detecting channels with specific properties to directly manipulate the latent code, which is limited by the entanglement of the latent space.
To detect the attribute-specific channels, we propose a novel detection method in the context of pre-trained classifiers. We analyse the gradients layer by layer on the style space. The intensities of the gradients indicate the channel's responses to specific attributes.
The latent style codes of channels control separate attributes in the layers. 
We choose channels with the top-$k$ gradients to control specific attributes in the maximum response layer. 
We implement single-channel and multi-channel manipulations with a certain attribute. 
Our methods can accurately detect relevant channels for a large number of face attributes. Extensive qualitative and quantitative results demonstrate that the proposed methods outperform state-of-the-art methods in generalization and scalability.
\end{abstract}

\begin{keyword}
Semantic Manipulation \sep Face Editing \sep Generative Adversarial Networks \sep StyleGAN
\end{keyword}

\end{frontmatter}



\section{Introduction}

Generative Adversarial Networks (GANs) can synthesize high-quality images from the latent space. Obtained from Gaussian independent sampling, the latent space $\mathcal{Z}$ is mapped to the image space by layer-wise convolutional upsampling ~\cite{karras2018progressive}. 
StyleGAN~\cite{karras2019style,karras2020analyzing} introduces the style codes to represent the different semantics, which transforms the latent space $\mathcal{Z}$ to the latent space $\mathcal{W}$ by the mapping network. StyleSpace~\cite{wu2021stylespace} decouples $\mathcal{W}$ to the style space $\mathcal{S}$ in each modulation layer.
The study of disentangled representations ~\cite{lee2018diverse} reveals the mapping structure of GANs from the latent space to image space. The disentangled semantic attributes are discovered along the specific direction in the $\mathcal{Z}$, $\mathcal{W}$ or $\mathcal{S}$ space. Manipulating the latent code has become a novel research topic to achieve controllable semantic editing of synthetic images in GANs.

The purpose of manipulating the latent code is to achieve the semantic editing of the generated images.
Detection of the attribute-specific channels is the key problem in the latent space.  Since the latent space $\mathcal{W}$ and $\mathcal{S}$ are more disentangled than the latent space $\mathcal{Z}$, most studies analyse attribute channels based on pretrained StyleGAN. 
Semantic attribute detection methods can be roughly divided into supervised approaches and unsupervised approaches. For unsupervised approaches, GANSpace\cite{harkonen2020ganspace} adopts PCA to analyse meaningful editing directions in the latent space $\mathcal{W}$. 
Collins \etal~\cite{2020_edit} achieve the semantic transformation about $\mathbf{w}$ by editing the related channels from source to target images.
SeFa~\cite{shen2021closed} analyses the affine transformation from $\mathcal{W}$ to $\mathcal{S}$ space by eigenvector decomposition. 
The biggest drawback of the unsupervised methods is that the editing direction needs to be interpreted by humans. These methods cannot pinpoint the channels of specific semantics. 

For supervised approaches, attribute annotations come from pre-trained classifiers or samples. Therefore, supervised methods can find the attribute-specific channels.
InterfaceGAN~\cite{shen2020interfacegan} finds the hyperplanes in the latent space $\mathcal{W}$ by the linear support vector machines (SVMs), which is pre-trained on the attribute annotations dataset. 
StyleRig~\cite{tewari2020stylerig} transforms pose, illumination, and expression attributes by reducing the distance between the generated image and the original image. 
Due to the entanglement of the latent space, those methods greatly modify other characteristics when transforming the specific attribute. 
In order to only modify the specified property, StyleFlow~\cite{abdal2021styleflow} adopts a normalizing flow network to establish a bi-directional mapping of latent codes and attributes.
Khodadadeh~\etal~\cite{9706683} embed the attributes to the latent code $w$ by fusing the transformation network. Those methods all edits on the $\mathcal{W}$ space. However, the transformation network needs to learn the mapping for the specific attribute editing in the latent space. It requires a large amount of semantic annotation samples. 
Furthermore, those implicit methods are less flexible when performing multi-attribute and continuous manipulation.
Even though they adopt a multi-task model for multi-attribute manipulation, the network fails to make continuous manipulation due to the lack of editable semantic directions. 
The style space $\mathcal{S}$ exhibits the independence of channels related to attributes. StyleSpace~\cite{wu2021stylespace} proposes the single channel manipulation for attribute regions. StyleSpace locates the attribute-specific channels by analysing the mean and variance on the positive samples. 

In this paper, we propose a new method to detect attribute channels based on gradients in the style space $\mathcal{S}$. Considering facial attribute editing, we obtain the gradient of $s$ in the context of pre-trained attribute classifiers.
The intensity of gradient indicates the channel's response to the attribute. 
We discover that the average gradients at layers are distinct for the specific attribute in the $\mathcal{S}$ space. 
Only a tiny fraction of the channels respond strongly to a certain attribute. 
Inspired by those observations, we choose the top-$k$ channels with the largest gradients in each layer. Manipulating these channels is related to editing the corresponding attributes. We apply this method to detect the channels about attribute regions and semantic attributes. Our method detects relevant channels on $10$ attribute regions and $40$ semantic attributes, which greatly surpasses the number of editable attributes with other methods. 
In the facial editing experiment, we also find that some attributes can be controlled through a single channel like ``lipstick" attribute, while the other attributes correspond to multiple channels like ``eyeglasses" attribute. 
Therefore, we propose the single-channel and multi-channel manipulation for various attributes. 
We quantify the editing strength of single-channel and multi-channel manipulations.
The unit editing intensity of a single channel is derived from the variance of the samples in the dataset. The editing direction of the multi-channel comes from the normalization of the average gradients in the positive samples.
To the best of our knowledge, our methods have great generalization on various attributes and can accurately locate specified attribute channels.
Our main contributions are summarized as follows:
\begin{itemize}
\item We propose a new detection method based on layer-wise gradients in the style space. We select the corresponding channels by the sorted gradients at each layer for a specific attribute.
Our method can detect a variety of facial attributes including the regions and semantic attributes.
\item With the proposed detection method, we achieve single-channel and multi-channel semantic editing on the $\mathcal{S}$ space. We also quantify the editing strength of single-channel and multi-channel manipulations. 
\item In the facial editing experiments, various attributes correspond to different layers and channels. Some attributes can be controlled through a single channel, while the other attributes correspond to multiple channels. 
\end{itemize}

\subsection{Related work}
StyleGAN/StyleGAN2~\cite{karras2019style,karras2020analyzing} generates synthetic images by transforming the latent codes layer by layer. In this paper, StyleGAN stands for StyleGAN2 for brevity.
StyleGAN is made of a mapping network and a synthesis network.
The original code $\mathbf{z}$ is entangled to represent the semantics of the generated image. A mapping network transforms $\mathbf{z}$ into $\mathbf{w}$ for separating semantic representation.
The code $\mathbf{w}$ modulates the style by the affine transformation in the synthesis network. $\mathbf{w+}$ comes from $\mathbf{w}$ by duplicated $18$ times for deconvolution upsampling from $4\times4$ to $1024\times1024$ resolution in each layer of the synthesis network.
$\mathbf{w+}$ can be separately modified in different resolution layers, which is studied in style mixing~\cite{karras2019style} and image inversion~\cite{richardson2021encoding,tov2021designing}. 
Furthermore, StyleSpace~\cite{wu2021stylespace} proposes the $\mathcal{S}$ space to combine the embeddings in the tRGB layer and deconvolution layer of each up-sampling module.
The tRGB layer converts feature maps to RGB images in each resolution layer.
StyleSpace evaluates the $\mathcal{Z}$, $\mathcal{W}$, $\mathcal{W+}$ and $\mathcal{S}$ spaces using the DCI (disentanglement / completeness / informativeness) metrics~\cite{eastwood2018framework}. StyleSpace demonstrates that the style space $\mathcal{S}$ is more disentangled than the other latent spaces of StyleGAN.	

Manipulating latent codes meticulously in StyleGAN can edit the semantic characteristics of synthetic images. To manipulate the code on the latent space $\mathcal{W}$, InterfaceGAN~\cite{abdal2019image2stylegan} trains a linear support vector machine (SVM) with attribute annotation images. 
Analysing the hyperplane in the SVM, the edit vector $\boldsymbol{\epsilon}$ is discovered to perturb the original $\mathbf{w}$ so that $\mathbf{w^*} = \mathbf{w}+\boldsymbol{\epsilon}$ crosses the classification boundary.
InterfaceGAN is limited in the number of editable attributes due to training the SVM and analysing the hyperplanes with the annotated attribute dataset.
StyleFlow~\cite{abdal2021styleflow} proposes a conditional continuous normalizing flow network for mapping the code $\mathbf{w}$ with specific attributes.  Hou \etal~\cite{Hou2022GuidedStyleAK} trains a transformation network to map the latent space $\mathcal{W}$ to a new space associated with the attributes. The transformation network needs to be trained for editing different attributes. 
GANSpace~\cite{harkonen2020ganspace} analyses the latent space $\mathcal{W}$ by PCA and identifies semantic editing directions manually. 
GANSpace is unable to discover the editing direction for a specific semantic attribute.
In summary, these methods can be divided into explicit and implicit manipulation methods on the latent space $\mathcal{W}$. As the entanglement of the $\mathbf{w}$ code, both approaches slightly modify the other attributes when editing the specific attribute.
 
Besides manipulation in the $\mathcal{W}$ space, SeFa~\cite{shen2021closed} analyses the affine transformation from $\mathcal{W}$ to $\mathcal{S}$ space by eigenvector decomposition. SeFa cannot find the attribute-specific directions, which manually annotates the editing directions related the semantic attributes. 
Since latent code in the $\mathcal{S}$ space is more spatially disentangled, 
StyleSpace only selects one channel of $\mathbf{s}$ to manipulate the relevant properties.
StyleSpace calculates the variance of the style code about the positive samples to locate channels for specific attributes. 
Wang \etal~\cite{wang2021attribute} proposes attribute-specific control units for multi-channels editing about the latent code $\mathbf{s}$. 
The single-channel manipulation aims to locate the associated channels about attributes. The multi-channel manipulation focuses on balancing the editing intensity of multiple channels, which maintains the other attributes as much as possible along the editing directions. 
Different from the above methods, Wang \etal~\cite{Wang2022CrossDomainAD} manipulate faces in arbitrary domains using human 3D Morphable Model (3DMM)~\cite{egger20203d}. Nichol \etal~\cite{nichol2021glide} explore diffusion models about text-conditional image synthesis and manipulation. 
In this paper, we focus on the latent space $\mathcal{S}$ and investigate the gradients for attribute-specific manipulation. 

\section{Method}
The generator $G(\cdot)$ of StyleGAN learns a mapping from a $512$-dimensional latent $\mathbf{z} \in  \mathcal{R}^{512}$ to a 1024x1024 resolution image.  $\mathbf{w} \in \mathcal{R}^{512}$ is obtained by the transformation network $T(\mathbf{z})$.
The generator $G(\cdot)$ consists of $18$ layers about style modulation modules. 
Duplicated $18$ times from $\mathbf{w}$, $\mathbf{w+} \in \mathcal{R}^{512 \times 18}$ is applied to each style modulation module. 
As a simple copy of $\mathbf{w}$, the code $\mathbf{w+}$ is entangled in different generation layers.
\cite{richardson2021encoding,tov2021designing} directly recode $\mathbf{w+}$ to decouple the various semantics in modulation modules.  The $\mathcal{W}+$ space expands the $\mathcal{W}$ space to enrich the representation of the latent code.
Furthermore, StyleSpace separates two independent style codes $\mathbf{s}^i$ and $\mathbf{s}^{i+1}$ which are associated with $\mathbf{w}^i$ in the $i$-th convolutional modulation layer. $\mathbf{s}^i$ and $\mathbf{s}^{i+1}$ are acquired by performing affine transformation $\mathbf{A}^i\mathbf{w}^i$ and $\mathbf{A}^{i+1}\mathbf{w}^i$ on the $i$-th generation layer. 
In summary, there are $26$ independent style codes for modulation layers. 
We list the details about the generator $G(\cdot)$ on different latent spaces in the Table.~\ref{struct} of \ref{generator}.

In this paper, we aim to detect the attributes in the $\mathcal{S}$ space and manipulate the related channels for specific semantics. 
First, we investigate the structure of latent space $\mathcal{S}$ about a certain attribute of the images on the pre-trained StyleGAN. 
We mainly study two types of attributes about faces.
One is the object related to facial region, such as hair, mouth, etc. 
The other is high-level semantic attribute, which is described by classifiers or samples such as young, goatee, etc. 
Detecting the attribute-specific channels on the latent space $\mathcal{S}$,
 $\mathbf{s}^i_j$ means the $j$-th channel in the $i$-th modulation layer. StyleSpace demonstrates that a single style channel $\mathbf{s}^i_j$ can represent a certain semantic attribute of the face. 
However, StyleSpace is inefficient to locate the channels with semantic attributes. StyleSpace also fails to detect some semantic attributes like ``old'' about faces.
We propose a gradient-based method to detect the style channels for attribute-specific manipulation to improve detection efficiency and localization accuracy.
Then, we quantitatively manipulate these channels to implement semantic editing of images in the latent style space.

\begin{figure*}
\subfigure[The distribute of the top-$500$ with respect to the gradient intensities.]
{
\includegraphics[scale=0.35]{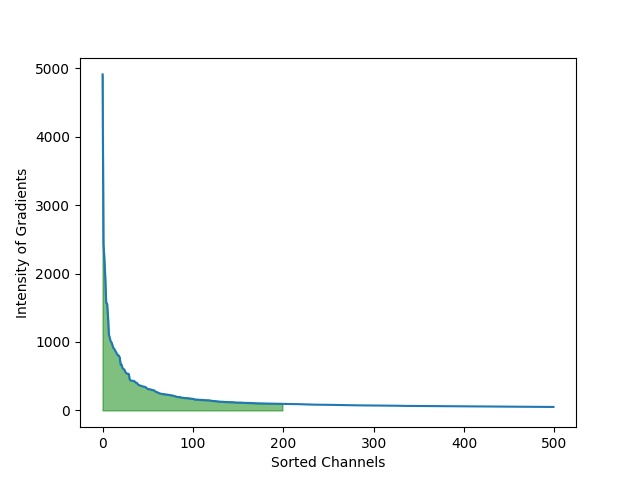}
\label{gradients}
}
\subfigure[The gradient intensities of different layer on the ``mouth'' region.]
{
\includegraphics[scale=0.35]{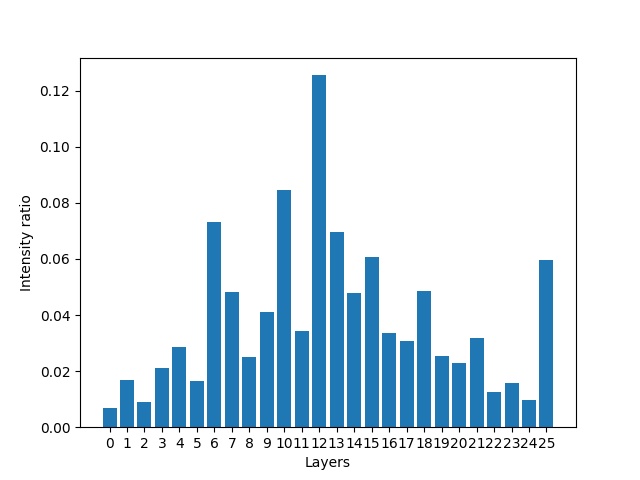}
\label{gradients_proportion}
}
\caption{Analyzation the gradients of channels and layers on the style space $\mathcal{S}$.}
\label{analyzations}
\end{figure*}

\begin{figure}
\begin{center}
\includegraphics[scale=0.5]{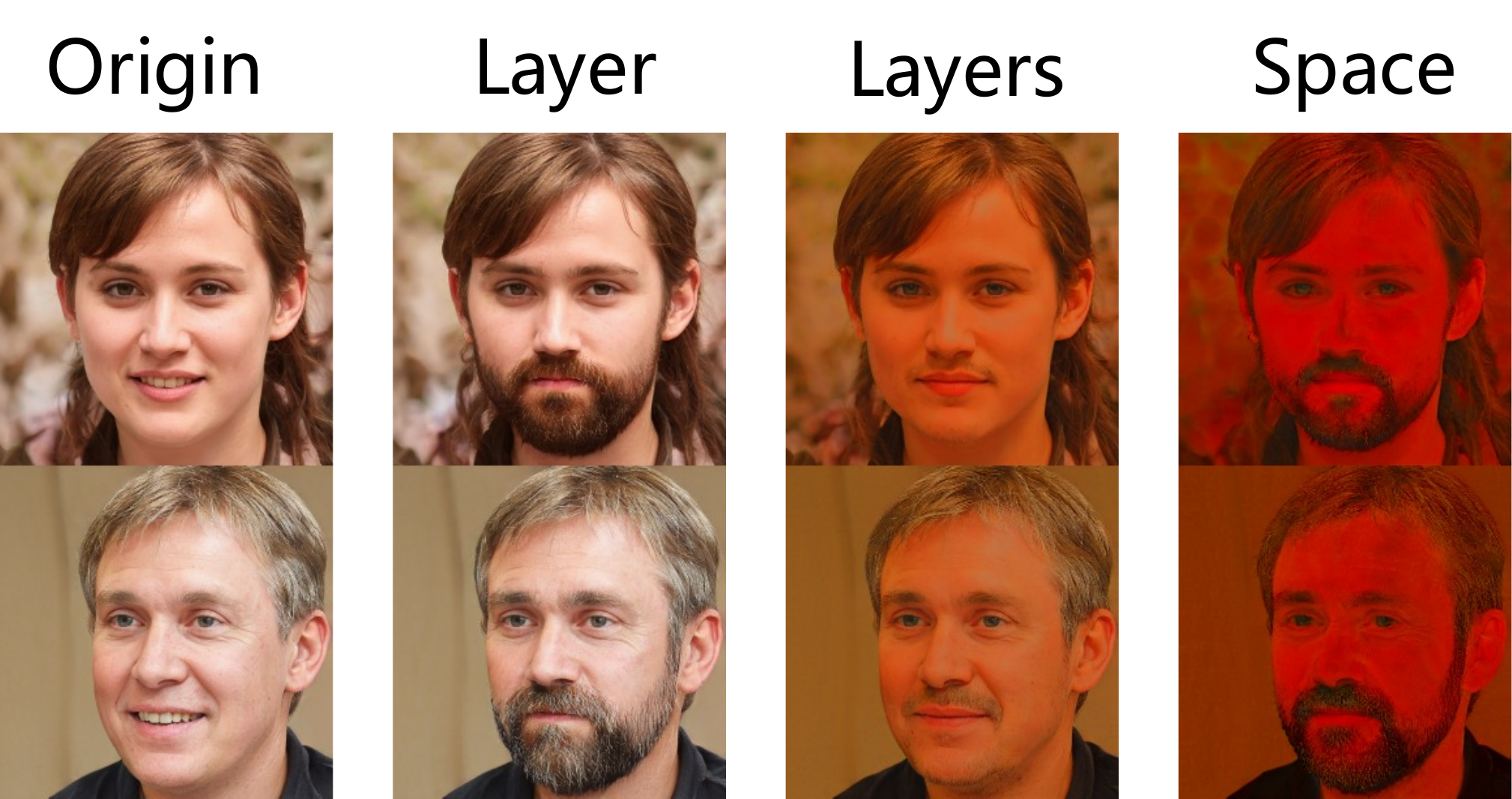}
\end{center}
\caption{Manipulation results about channel and layer.}
\label{images_proportion}
\end{figure}

\subsection{Analysing gradients with attribute regions}
First, we analyse the gradients of style channels that control the local visual semantic regions. Suppose the image $\mathbf{I}$ is generated from the $G(\mathbf{s})$. The latent code $\mathbf{s}$ corresponds to the internal representation of $\mathbf{w}$ in $G(\cdot)$.
A pre-trained image segmentation network is used to predict the semantic mask $\mathbf{M}$ of the generated image $\mathbf{I}$. The local semantic region $r$ is represented as $\mathbf{M}^r$. We compute the average gradient about the local semantic region $r$ with respect to each channel of $\mathbf{s}$ as follows,
\begin{equation}
\mathbf{g}_\mathbf{s}^r = \frac{1}{\sum_{p \in \mathbf{M}^r}}\sum_{p \in \mathbf{M}^r} \frac{\partial \mathbf{I}_p}{\partial \mathbf{s}},
\end{equation}
where $p$ denotes the index of each pixel in the semantic region $r$. 
The intensity of $\mathbf{g}_{s^{i}_{j}}^r$ denotes the response to the region $r$ about the $j$-th channel in the $i$-th layer. Fig.~\ref{gradients} shows the top-$500$ intensities of $\mathbf{g}_\mathbf{s}^r$ on the ``mouth" region. The distribution of $abs(\mathbf{g}_\mathbf{s}^r)$ is similar to the long-tailed distribution. The intensity of the gradient indicates the channel's response to a certain attribute region. 
The specific attribute is controlled by only a small number of channels, which are embedded in the subspace of the style space $\mathcal{S}$.
For the ``mouth" region, the sum of the top $200$ gradient strengths accounts for $40.6$\%, and most of the rest channels are inactive in the flatten StyleSpace $\mathcal{SF} \in \mathcal{R}^{9088}$. 
To locate the generation layer that has the greatest impact on the attribute region, Fig.~\ref{gradients_proportion} demonstrates the average gradient for each layer. 
We choose the top-$3$ layers to detect the channels for specific attributes.
First, we manually analyse the semantic of each channel for the attribute region. Before manipulating single channel editing, we calculate the mean and standard deviation $
\boldsymbol{\delta}$ of $1000$ samples to set the unit editing intensity for each style channel. 
We use standard deviation as the editing unit to achieve continuous manipulation on the style space $\mathcal{S}$ as follows
\begin{equation}
\mathbf{s}^* = \mathbf{s} + \alpha \boldsymbol{\epsilon}^*,
\end{equation}
where $\alpha$ determines the disturbance distance along the $\boldsymbol{\epsilon}^*$ unit direction.
$\boldsymbol{\epsilon}^*$ represents the unit editing direction of single channel, which is derived from standard deviation $\boldsymbol{\delta}$.
The manipulation results are visualized in Fig.~\ref{u_lip_activate} of \ref{gradients_topk}. 
Compared with Fig.~\ref{layers_activate} in ~\ref{hier}, we choose the top-$k$ channels based on the intensity of gradients, which more accurately locate layers and channels for specific attributes. Without data analysis on the sample dataset, we achieve the single channel editing based on the gradients of style code $\mathbf{s}$.
However, it is still significant to investigate the average gradients of the sample dataset. As shown in Fig.~\ref{hair_ayers_activate} of \ref{hier}, a series of style channels control the various hair attributes. For one style code $\mathbf{s}$ about generated image, the top-$k$ gradients only discover a subset of features in the the attribute region. For example, our method can detect the channel response to black hair via the synthetic image with black hair, but fails to detect the channel response to red hair. The average gradients over the test samples increases the number of detection channels on attribute regions.

Besides, for the detection channels, it is difficult for them to distinguish various semantics in the same attribute region. 
The different semantics are associated with the attribute region in the figures of \ref{gradients_topk} and~\ref{hier}. 
Channels in the same layer can jointly control attribute characteristics. 
Inspired by the observations, we select the top-$k$ channels of the same layer as a group to achieve multi-channel editing.
As Wang~\etal~\cite{wang2021attribute} pointed out, single-channel editing suffer from insufficient modification for complex semantics. Multi-channel editing focus on the fine-grained image manipulation on the style space. 
Our method selects the relevant channels in charge of specific attributes by gradient intensity about layers and channels. 
Fig.~\ref{images_proportion} shows the different manipulation results about the beard in the mouth region. In the second figure of Fig.~\ref{images_proportion}, we edit the original style code along the gradient direction of the top-$3$ channels on the $12$-th layer.
For the third figure, we normalize the gradients on the top-$3$ layers as $\boldsymbol{\epsilon}^*$ in the style space.
The last result is severely distorted by editing along the unit gradient direction on the style space. 
As shown in Fig.~\ref{images_proportion}, it is critical to filter suitable channels for specific attribute manipulation. 
Balancing the editing intensity of multi-channels is an open problem.
We'll dive into more techniques for multi-channel editing in the next subsection.

\subsection{Analysing gradients with semantic attributes}
The main disadvantage of the attribute region is the lack of described characteristics in the area. The above method cannot distinguish the shape, color, or other semantic features, which are associated with the same attribute region. Therefore, we extend the above method to the semantic attributes described by classifiers or samples. 
Positive and negative samples are annotated to implicitly represent semantic attributes.
Suppose the classifier model $F(\cdot)$ about binary semantic attribute is pretrained on the related attribute dataset. For the generated image $\mathbf{I}$, the gradients of the style code $\mathbf{s}$ are calculated for the classifier's predictions by the chain rule,
\begin{equation}
\mathbf{g}^a_\mathbf{s} = \frac{\partial F(\mathbf{I})}{\partial \mathbf{s}} = \frac{\partial F(\mathbf{I})}{\partial \mathbf{I}} \frac{\partial \mathbf{I}}{\partial \mathbf{s}}.
\end{equation}
Similar to the method on attribute regions, we sort the gradients $
\mathbf{g}^a_\mathbf{s}$ and choose the attribute-specific channels for single-channel manipulations. For single style code $\mathbf{s}$, the manipulation results by sorting the gradient $\mathbf{g}^a_\mathbf{s}$ can not be interpretable to the specified attributes. 
To discover the channels related to the specified features, we investigate the average gradients $\mathbf{g}^a_\mathbf{s}$ on the attribute dataset. For the negative samples, since the generated images do not contain the specified semantic attributes, they cannot locate the relevant style channels through the gradient $\mathbf{g}^a_\mathbf{s}$. 
We focus on the style codes whose generated images are classified as positive samples by a specific attribute classifier. 

The gradients are applied to detect the channels for semantic attributes on the positive samples. 
$\mathbf{s}_i$ denotes the $i$-the style code in the set $P = \{\mathbf{s}_1, \mathbf{s}_2, ..., \mathbf{s}_n\}$. Each $\mathbf{s}^i$ is converted from $\mathbf{z}_i$ by the mapping network of StyleGAN.
$P$ represents the style code set about positive samples which are generated from generator $G(\mathbf{z}_i)$ and labelled by the classifier $F(\mathbf{I}_i)$. 
We calculate the average of the gradients about positive samples on $P$ as 
\begin{equation}
Average(\mathbf{g}^a_{\mathbf{s}_i})_{\mathbf{s}_i \in P} = \frac{1}{|P|}\sum_{\mathbf{s}_i \in P} \mathbf{g}^a_{\mathbf{s}_i}.
\end{equation}
We analyse the single-channel manipulation in the top-$k$ channels of sorted $Average(\mathbf{g}^a_{\mathbf{s}_i})_{\mathbf{s}_i \in P}$. 
In Fig.~\ref{young_activate} of \ref{gradients_topk}, we detect the single channel ``3-435" which controls the skin wrinkles to modulate ``young" attribute. The channel ``3-435" means the $435$-th channel in $3$-th layer.
Our method finds the channels for the ``young" attribute in the style space, where the responding channels are not detected by the StyleSpace method.
Our method can detect the attribute-specific channels with $20$-$40$ positive samples.

Manipulation with a single channel has some drawbacks. 
First, a single-style channel may be incomplete when representing an attribute. 
As shown in Fig.~\ref{young_activate} of \ref{gradients_topk}, the attribute ``young’' is not only related to the smoothness of the skin, but also to the hair, eyebrows, etc. 
Single-channel manipulation ignores the representation of other channels. 
As the analysis of gradients in Fig.~\ref{analyzations}, a tiny fraction of the channels are associated with the attribute properties in the style space. 
Therefore, we extend the single-channel manipulation toward multi-channel manipulation. 
Since style codes control similar semantic attributes in each layer, we select the top-$k$ channels of the same layer as an editing unit. 
$C^k_l$ is a set of the top-$k$ channels on the $l$-th style layer, i.e.,
\begin{equation}
C^k_l = \{\text{Top-}k \; \text{channels on }Average(\mathbf{g}^a_{\mathbf{s_i},l})_{\mathbf{s_i} \in P}\}.
\end{equation}
As shown in the Fig.~\ref{black_12} and Fig.~\ref{black_6} of \ref{gradients_topk}, we visualize the manipulation result about ``hair'' attribute on the $12$-th  and $6$-th layers. Channels in $C^k_l$ tend to generate similar semantics in the $k$-th layer. The unit editing direction for multi-channel manipulation is undefined, which is different from the unit editing distance for single-channel manipulation. 
We set the editing direction on the $C^k_l$ channels according to the average gradients. 
The unit editing direction of the multi-channel manipulation is 
\begin{equation}
\boldsymbol{\epsilon_m} = Average(\mathbf{g}^a_{\mathbf{s}_i,C^k_l})_{\mathbf{s_i} \in P}.
\end{equation}
By selecting the appropriate style layers and the number of channels, we achieve multi-channel manipulation with fine-grained semantic attribute on the style space $\mathcal{S}$.

\section{Experiment}
\subsection{Experiment setup}
We evaluate the proposed methods on the facial attribute editing tasks. 
StyleGAN is pre-trained on the Flickr-Faces-HQ Dataset (FFHQ)~\cite{karras2019style}. 
We randomly sample and generate 1000 synthetic images with the latent code $\mathbf{z}$ and $\mathbf{s}$ via StyleGAN.
This dataset is adopted for semantic attributes extraction and manipulation of images, which is smaller than other methods.
As a semantic segmentation model, BiSeNet~\cite{2018_bisenet} is used to segment basic attributes about faces such as eyes, hair, etc. We also train the classifiers for the $40$ binary attributes in CelebA\cite{liu2015faceattributes}.  We adopt ResNet50~\cite{2016_resnet} as backbone to build facial attribute classifiers. 

Following StyleSpace, we adopt Attribute Dependency (AD)~\cite{wu2021stylespace} as metric to measure the semantic disentanglement of manipulated results. 
Attribute dependency computes the attribute changes of the logits between the manipulation result and the original image. 
The logits of the $i$-th classifier is denoted as $l_{i}$.  $\Delta l_{i}$ measures the change of logits between the original and the manipulated images. $\sigma(l_i)$ means the standard deviation about $l_i$.
We denote the $\mathcal{A}$ represents the set of editable attributes. 
The $AD_t$ computes the editing discrepancy in the target attribute $t$.
The $AD_o$ is evaluated the average change excluding the editing attribute $t$ on the positive samples, which defined as $E\left(\frac{1}{k} \sum_{i \in \mathcal{A} \backslash t}\left(\frac{\Delta l_{i}}{\sigma\left(l_{i}\right)}\right)\right)$, where $k=|\mathcal{A}| - 1$.
However, $AD_t$ and $AD_o$ are not only related to the editing direction, but also to the editing strength along that direction. Due to the inconsistency of unit edit distances, we can not directly compare $AD_t$ and $AD_o$ metrics with different editing methods. We propose to adopt the ratio of $AD_t$ and $AD_o$ to address the problem of editing intensity.
The ratio $\frac{AD_t}{AD_o}$ indicates the independence of the editing direction about attributes. 
Evaluation metrics are computed on positive samples from the test dataset. Each attribute contains roughly $20$-$50$ positive samples.

\subsection{The manipulation results based on gradients}

\subsubsection{Results of local attribute regions}
\begin{figure}
\begin{center}
\includegraphics[scale=0.24]{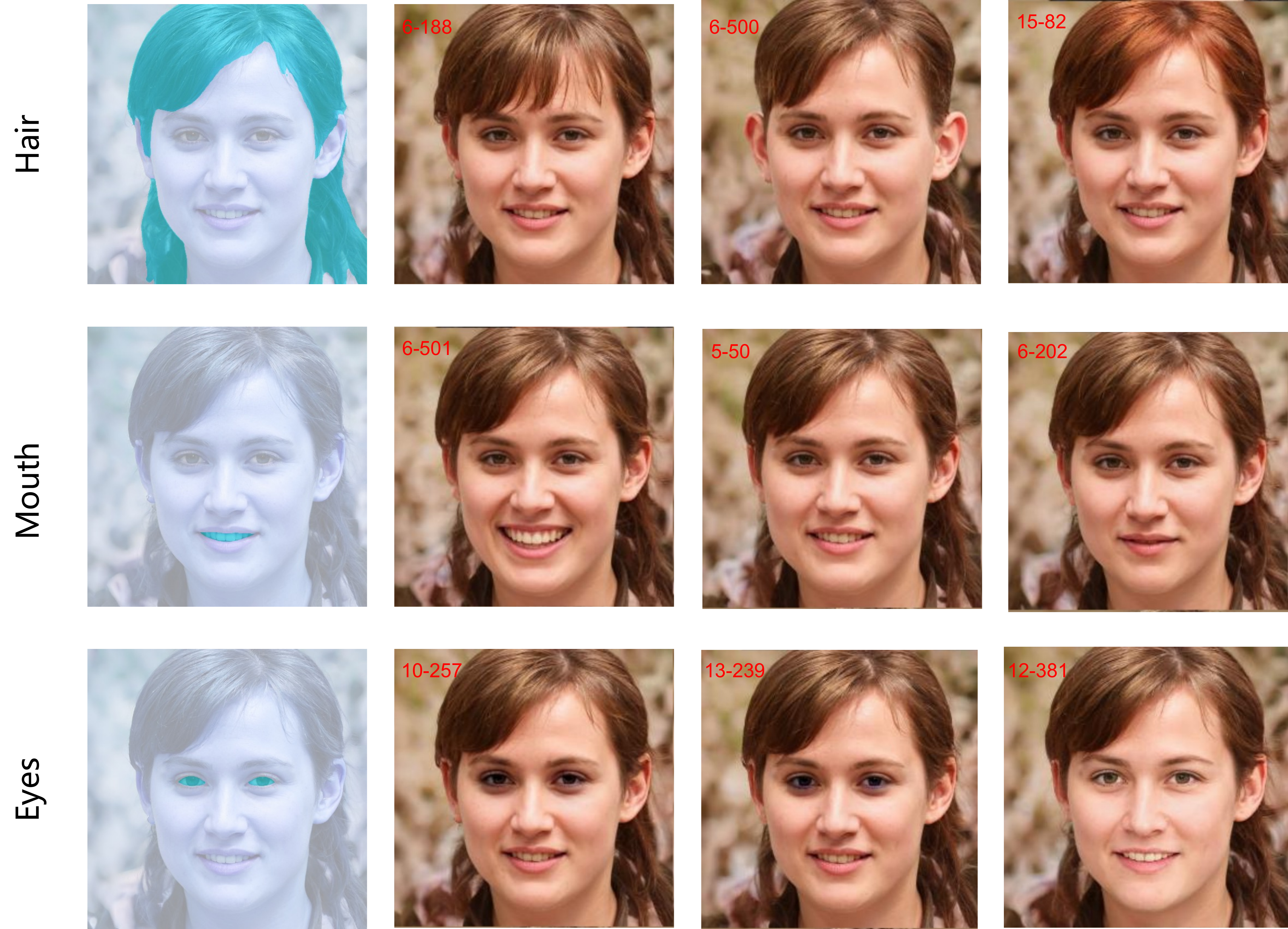}
\caption{Examples of manipulation on attribute regions.}
\label{manipuate_regions}
\end{center}
\end{figure}

Fig.~\ref{manipuate_regions} shows the manipulation results of detection channels on the attribute region. The attribute area contains a variety of semantic attributes. As shown in Fig.~\ref{manipuate_regions}, our method detects the channels about region of ``hair'' including the shape and color properties. 
For a single image, we manually select the channels associated with a certain attribute from the top-$10$ about intensity of gradients. 
This method roughly detects the generated region of each channel but fails to locate the responding channels for specific semantic attributes.
Furthermore, since the method ignores the influences outside the attribute region, the detection channels may control specific attribute area as well as other attribute areas. 
 
\begin{figure*}
\begin{center}
\includegraphics[scale=0.40]{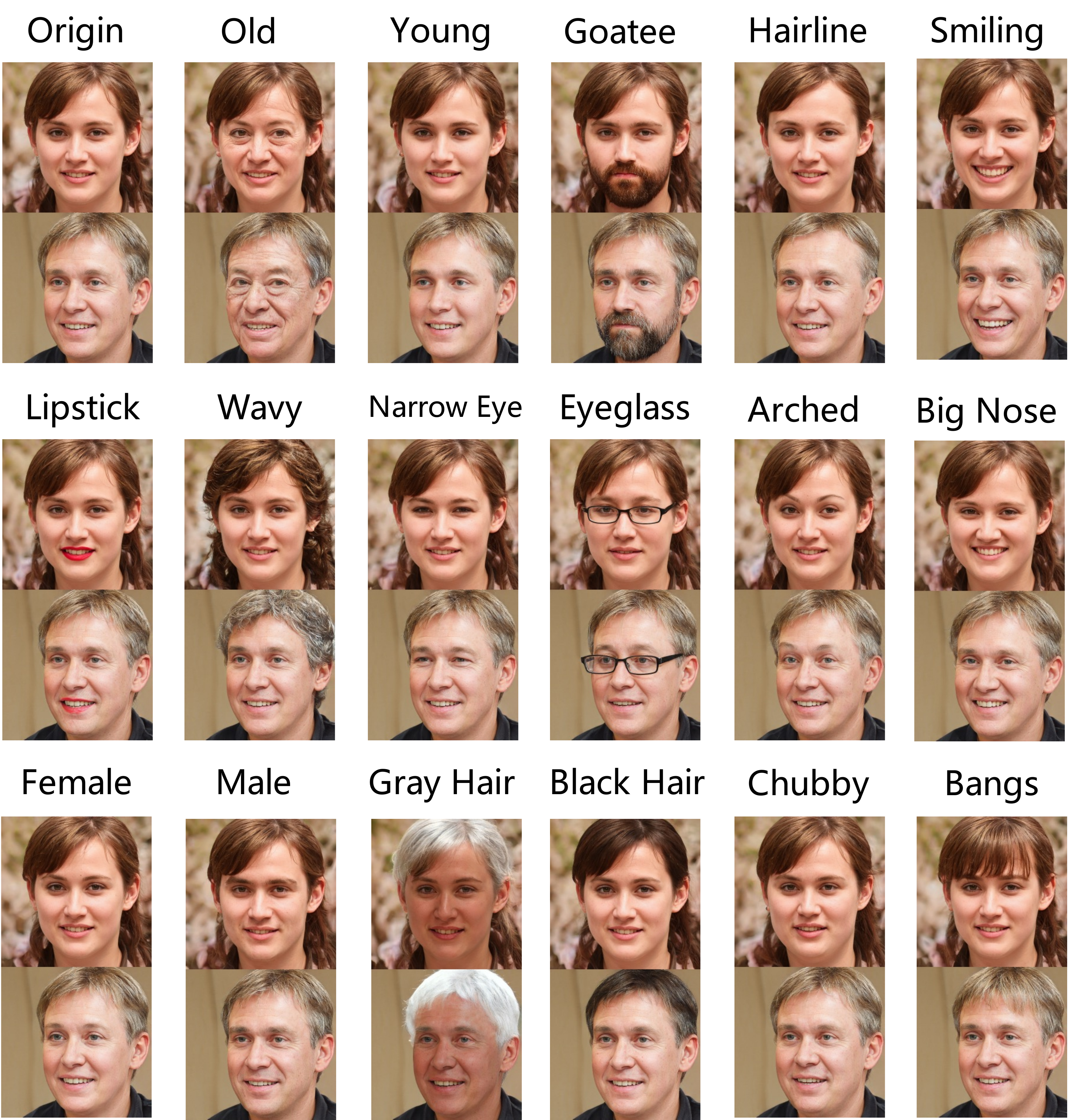}
\caption{Examples of manipulation on semantic attributes.}
\label{manipuate_samples}
\end{center}
\end{figure*}

\begin{landscape}
\begin{table*}
\caption{The style codes about layers and channels correspond to semantic attributes.}
\label{table_eemantic}
\begin{tabular}{c|ccccccc}
\hline
\hline
Attribute & Old / Young & Goatee& Hairline& Smiling & Lipstick & Wavy Hair & Narrow Eye\\
\hline
(layer, channel, rank) & 
\makecell{(9, 435, 1)}&
\makecell{(9, 421, 1)\\(9, 6, 2)\\(12,237,3)} & 
\makecell{(6, 322, 1)\\(6, 504, 2)\\(6, 364, 3)}& 
\makecell{(6, 501, 1)\\(6, 113, 2)\\(6, 378, 4)}&
\makecell{(15, 45, 1)}&
\makecell{(9, 475, 1)\\(6, 323, 1)\\(6, 500, 3)}&
\makecell{(11, 257, 1)\\(9, 63, 2)\\(14, 239, 3)}
\\
\hline
\hline
Attribute & Eyeglass & Arched &  Big Nose & Female / Male & Colour Hair  &Chubby &Bangs \\ 
\hline
(layer, channel, rank) &
\makecell{(3, 228, 1)\\(3, 120, 2)\\(2, 175, 3)}&
\makecell{(6, 35, 1)\\(9, 340, 2)}&
\makecell{(6, 501, 1)\\(6, 110, 2)}&
\makecell{(9, 6, 1)}&
\makecell{(11, 286, 1)\\(12, 424, 1)\\(15, 62, 1)}&
\makecell{(6, 113, 1)\\(6, 378, 2)\\(6, 104, 3)}&
\makecell{(6, 188, 1)\\(6, 322, 2)\\(5, 414, 3)}
\\
\hline
\hline

\end{tabular}
\end{table*}

\begin{table*}
\caption{Comparison of quantitative results measured by using different metrics. }
\label{table_metrics}
\begin{tabular}{c|ccccc}
\hline
\multirow{2}*{Methods} & Eyeglasses & Goatee & Smiling & Gender & Black Hair \\
~ & \makecell{($AD_t$, $AD_o$, Ratio)} & \makecell{($AD_t$, $AD_o$, Ratio)} & \makecell{($AD_t$, $AD_o$, Ratio)}  &\makecell{($AD_t$, $AD_o$, Ratio)}  &\makecell{($AD_t$, $AD_o$, Ratio)} \\
 \hline
StyleSpace & (0.40, 0.76, 0.53) & (2.76, 0.50, 5.49) & (2.94, 1.19, 2.46) & (2.54, 1.34. 1.89) & (4.38, 0.54, 8.03) \\
Single-channel &(2.61, 0.35, 7.31) & (2,36, 0.38, \textbf{6.20}) & (5.67, 0.67, 8.42) & (5.08, 1.12, \textbf{4.54}) & (4.38, 0.54, 8.03)\\
Multi-channel & (5.28, 0.71, \textbf{7.4}) & (6.72, 1.19, 3.50) & (7.80, 0.88, \textbf{8.88}) & (5.95, 1.36, 4.39) & (10.05, 1.04, \textbf{9.61})\\
\hline
\end{tabular}
\end{table*}

\end{landscape}

\subsubsection{Results of semantic attributes}
\begin{figure*}
\includegraphics[scale=0.30]{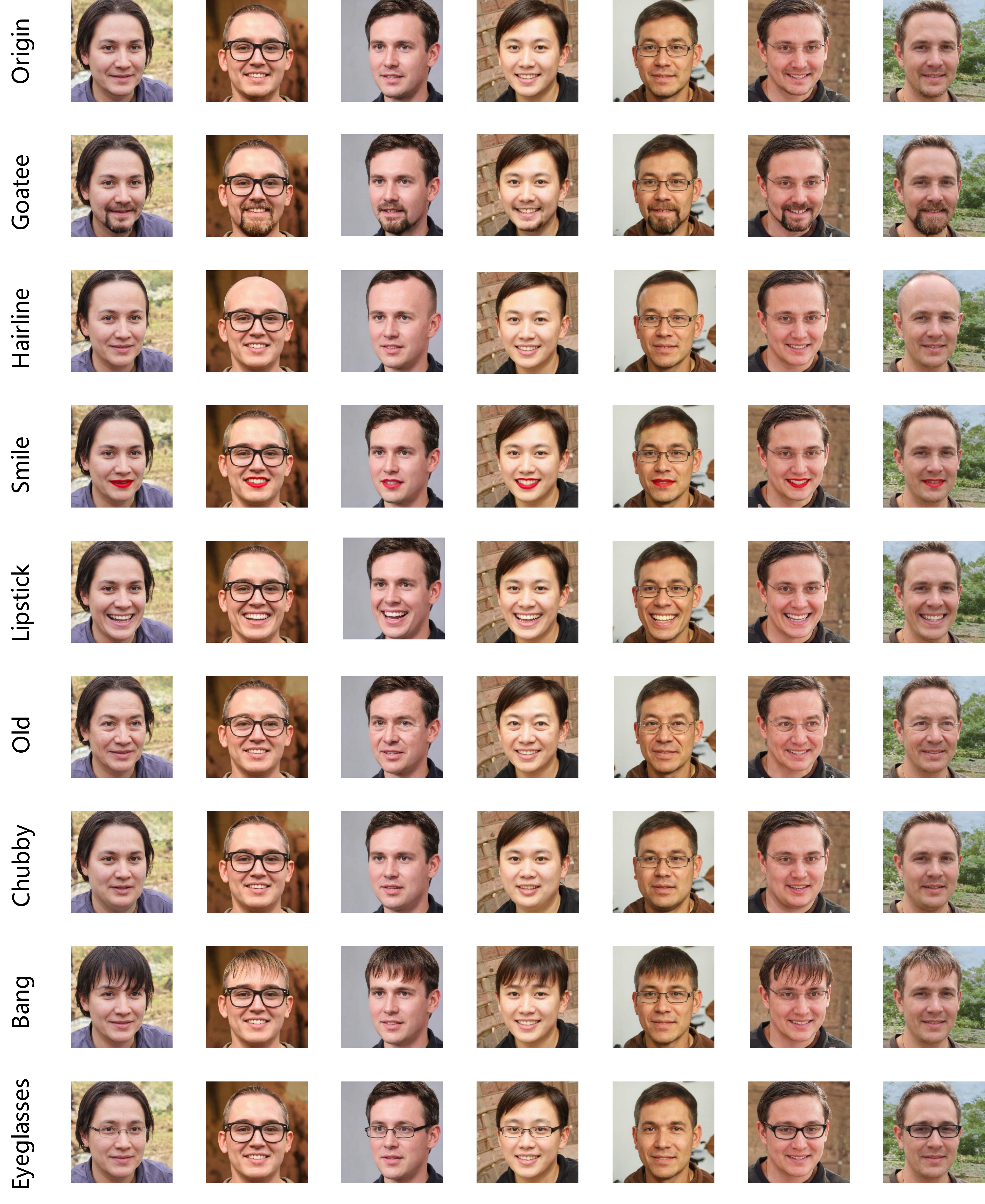}
\caption{Manipulation with various specific semantic attributes on the real images.}
\label{other_samples}
\end{figure*}

Fig.~\ref{manipuate_samples} demonstrates the editing results of  various attribute-specific semantics on the same sample.
Our method exploits the hierarchical gradients of $\mathbf{s}$ in each layer to detect different semantic semantics including the shapes, colors, and abstract features. 
Our manipulation approaches have great generalization which achieve semantic-specific editing on about $40$ facial attributes. 
Each detection channel can control consistent property for different synthesis images.
As shown in Fig.~\ref{other_samples}, the detection channels for the specific attributes are transferred to the real image manipulations by inversion methods~\cite{tov2021designing}.
Table.~\ref{table_eemantic} lists the corresponding layers and channels for the attribute-specific manipulations. We also investigate the single-channel and multi-channel manipulation for each semantic attribute. 
As shown in Table.~\ref{table_eemantic}, the single-channel manipulation is enough for simple attribute like ``lipstick" as the disentangled representation in the style space. However, a series of channels are detected for the complex attributes. 
Each responding channel controls the part of the attribute. 
The multi-channel manipulation implements fine-grained editing for $\mathbf{s}$.


\subsection{Comparison with other method}
We compare our approaches with other manipulation methods. 
Table.~\ref{table_metrics} demonstrates the $AD_t$, $AD_o$ and radio metrics about StyleSpace and our methods in the style space. For the ``eyeglasses" attribute, the radio of StyleSpace is less $1$, which means that it fails to detect the relevant channels. Our proposed methods edit the latent codes $\mathbf{s}$ for specific properties with slightly altering other attributes. Compared to single-channel manipulation, the multi-channel manipulation method achieves a larger editing distance while maintaining radio $\frac{AD_t}{AD_o}$.
As a single attribute is detected in the same style space, the detection channels have overlaps between our method and StyleSpace.  Both our method and StyleSpace locate the ``15-45'' channel about the ``lipstick" attribute. 
However, compared with StyleSpace, our method has a better generalization. Our method detects more editable attributes which are controlled by single-channel and multi-channel. For example, StyleSpace fails to detect the disentangled channels, which control for the ``young'' and ``eyeglasses'' attributes. 
Furthermore, StyleSpace computes the Jacobian matrix of the style code $\mathbf{s}$ for the attribute region, the detection time is linear to the number of pixels in the image. 
The detection time of our method is less than that of StyleSpace, even though StyleSpace downscales the image to $32 \times 32$ to reduce the computation of Jacobian.

\begin{figure}
\begin{center}
\includegraphics[scale=0.45]{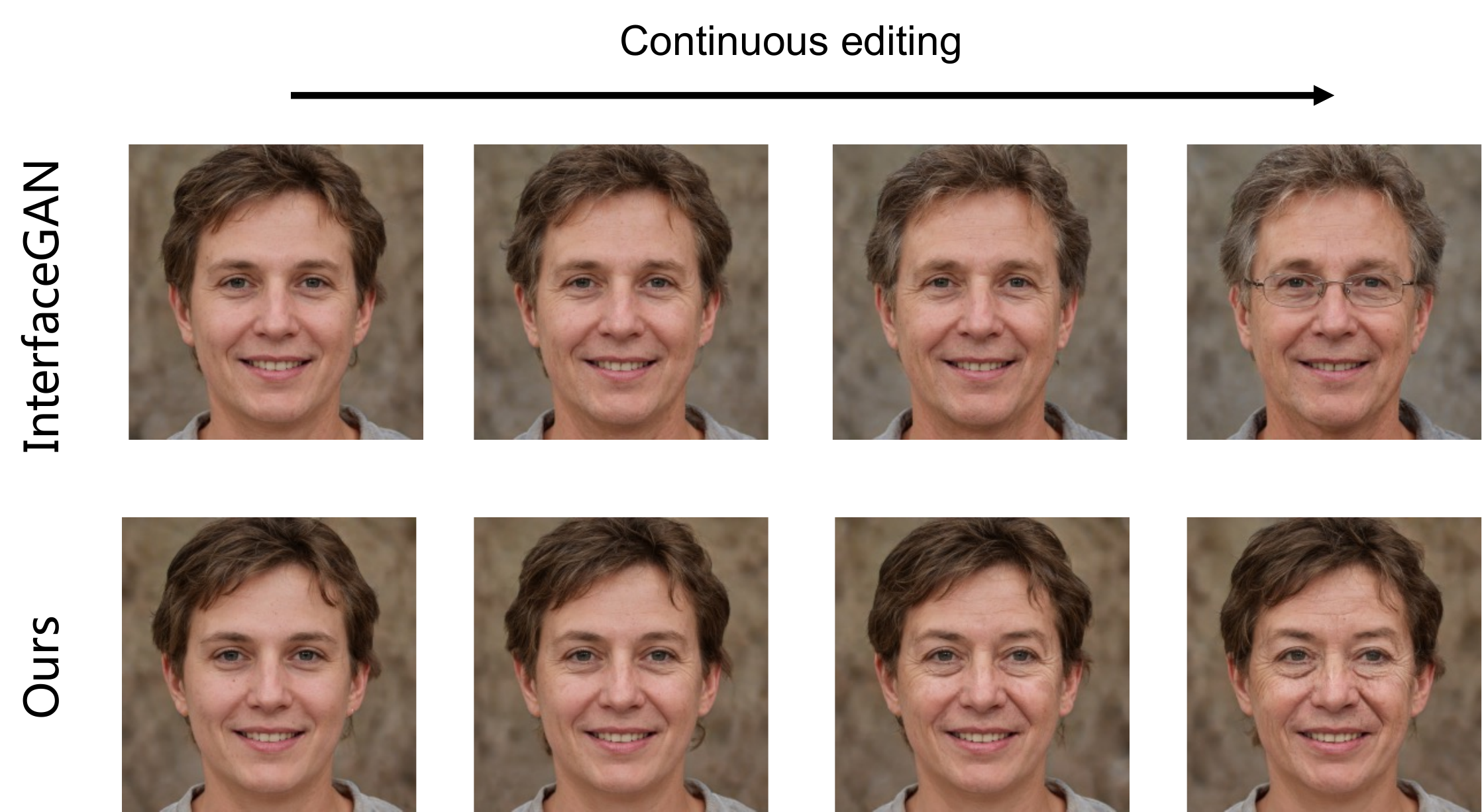}
\caption{Comparison the continuous manipulation results of ``old'' attribute about InterfaceGAN and our methods.}
\label{interface}
\end{center}
\end{figure}

We visualize the manipulation results about InterfaceGAN~\cite{shen2020interfacegan} and ours in the $\mathcal{W}$ and $\mathcal{S}$ space in Fig.~\ref{interface}. We apply the continuous manipulation along the unit editing direction. As shown in Fig.~\ref{interface}, the editing direction of InterfaceGAN is entangled which influence the ``gender'' and ``eyeglasses" attributes when increasing the editing intensity of ``old" attribute. 
The identity information of the original image may migrate with the image editing intensity. 
Our method shows the consistency of manipulation on attribute-specific channels while maintaining other properties.

Besides explicitly manipulating on latent spaces, some methods are also investigated to implicitly fuse specified attributes and latent codes. Without detecting the attribute-specific channels,  StyleFlow~\cite{abdal2021styleflow} and Lang \etal~\cite{Lang2021ExplainingIS} train a transformation network to combine the latent code with the semantic attributes. The transformation network not only needs to be trained on large semantic annotation datasets, but also limits the number of edited attributes. Siavash \etal~\cite{9706683} generate the $400$K samples whose attributes are extracted by the Microsoft Face API\footnote{https://azure.microsoft.com/en-in/services/ cognitive-services/face}. 
Without the generation and annotation on images, our method integrates the other attribute classifiers end-to-end to manipulate the attribute-specific editing.
Furthermore, explicit manipulation methods are more flexible than implicit ones. Explicit manipulation of the latent code achieves continuous editing by adjusting the intensity of editing. 
Table~\ref{compare_table} summaries and qualitatively evaluates various approaches from multiple perspectives.

\begin{table*}
\caption{Comparing manipulation methods from multiple perspectives.}
\label{compare_table}
\begin{center}
\begin{tabular}{c|cccccc}
\hline
Method & Model & Samples & Editable attributes & Independence \\
\hline
InterFaceGAN~\cite{shen2020interfacegan} & SVMs & 10K & 5 & Low \\ 
StyleSpace~\cite{wu2021stylespace} & Classifiers & 10-50 & 26 & High \\
Wang \etal~\cite{wang2021attribute} & Classifiers & 10K & 30 & High \\
StyleFlow~\cite{abdal2021styleflow} & Network & 10k+& 5 & Medium \\
Hou \etal~\cite{Hou2022GuidedStyleAK} & Network & 200K+ & 15 &Medium \\
Siavash \etal~\cite{9706683} & Network &400K+ & 35 & Medium \\
Our & Classifiers & 10-50 & 35 & High \\
\hline
\end{tabular}
\end{center}
\end{table*}

\subsection{Discussion and analysis}
\begin{figure}
\begin{center}
\includegraphics[scale=0.3]{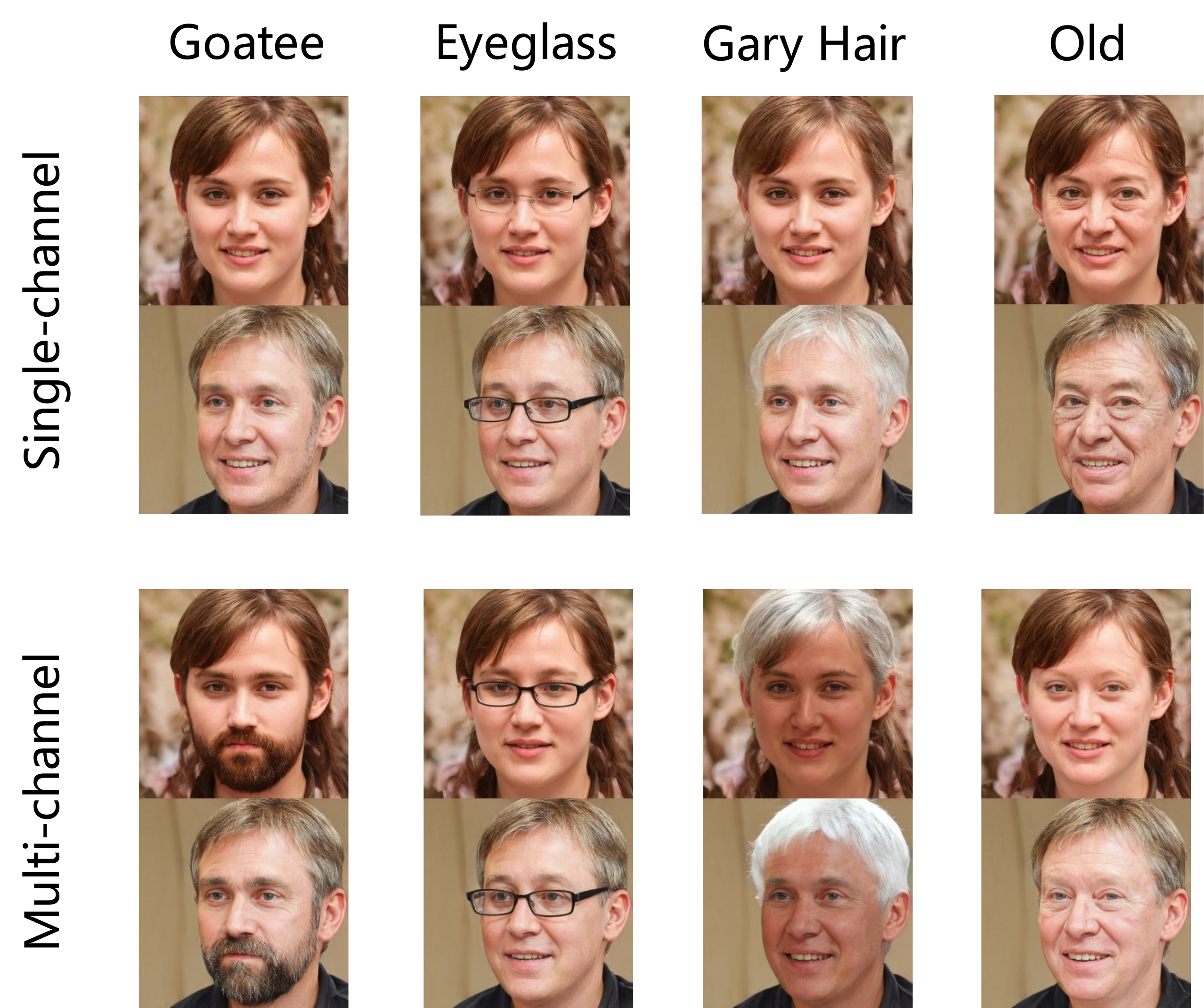}
\caption{Comparative examples on single-channel and multi-channel manipulation.}
\label{manipuate_single}
\end{center}
\end{figure}

\textbf{Single-channel and multi-channel manipulation:} Each channel controls a specific attribute in the style space. For a image, only a small fraction of the channels have significant influences on generated images according to the detection of gradients. For a particular semantic attribute, some channels distributing the different layers are responsible for modulating its style. As shown in Fig.~\ref{manipuate_single}, single-channel manipulation is insufficient to modulate entire semantic attributes. Multi-channel manipulation makes up the drawbacks of single-channel manipulation.
However, when editing multiple channels, choosing the number of channels and editing intensity is not a trivial task. We simply normalize the top-$k$ average gradients as the unit editing direction. Sometimes, it fails to generate reliable results such as the editing of ``gray hair" in Fig.~\ref{manipuate_single}.
Due to the problem of editing parameters, multi-channel manipulation is not necessary when the single-channel manipulation suffices to edit specific attributes.

\textbf{Style editing about layers:} Since the style modification of the high-level layer involves the color of the whole image or attribute regions, we filter the styles of $16$-$26$ layers in the previous experiments. 
As shown in Fig.~\ref{gradients_proportion}, we focus on the gradients of the channels to detect semantic attributes, especially at layers $6$, $10$ and $12$. 
Manipulating the channel of the high-level layer will distort the image, like the results about the ``$18$-$20$'' channel in Fig.~\ref{u_lip_activate} of \ref{gradients_topk}. We first select the control layer and then choose the top-$k$ channels by the average gradient of layers. We also found that above the $15$-th layer, the channels cannot modify the shape of the attribute and the slight modification to them affects the overall image color. Therefore, we filter out layers and channels above the $15$-th layer.

\textbf{Continuous editing about the intensity:}
The unit intensity of single-channel manipulation is derived from the standard deviation $\boldsymbol{\delta}$, which is calculated in the test dataset for each channel. After normalizing the channels, the manipulation intensity of a single channel is limited to [$-3\boldsymbol{\delta}$, $3\boldsymbol{\delta}$] according to Pauta criterion~\cite{li2016outlier}. In the continuous editing of single-channel manipulation, we not only detect the edited channels, but also determine the range of editing intensity.
However, as for the multi-channel manipulation, we are concerned with the balance of multi-channel manipulation. We simply normalize the top-$k$ average gradients as the unit editing direction. We adopt continuous editing to determine the range of editing strengths.


\section{Conclusion}
In this paper, we analyse the gradients about the attribute regions and semantic properties on the style space of StyleGAN. Only a tiny fraction of the channels are associated with the attribute properties in the style space. 
With the pre-trained attribute classifiers, we detect the corresponding channels for specific attributes end-to-end  by ranking gradient strengths.
Our method can quickly locate the channels on facial semantic attributes with $20$-$50$ positive samples. 
We also apply the single-channel and multi-channel methods for attribute-specific semantic manipulation on the style space.
Our approaches have great generalization to achieve semantic editing on about $35$ facial attributes. 
Our work explores the semantic interpretation of hidden representations on the style space of StyleGAN.

\section{Acknowledgement}
This work was supported in part by the National Natural Science Foundation of China under Grant 61876076.

\bibliography{elsarticle-template}

\appendix
\section{Structure of StyleGAN in the hidden spaces}
\label{generator}
Table ~\ref{struct} lists the structure of latent codes in the $\mathcal{W}+$ and $\mathcal{S}$ spaces for StyleGAN.
Each resolution module in generator $G(\cdot)$ consists of two convolutional layers for feature map and one convolutional layer that converts the second feature map to an RGB image (referred to tRGB layer). There are $17$ layers about $\mathbf{w}$ and $26$ layers about $\mathbf{s}$. The size of all $\mathbf{w^i}$ vectors is $512$. 
$\mathbf{s^i}$ can be regarded as the feature of $\mathbf{w^i}$  which is embedded in the linear space by $\mathbf{A}^i$. The $\mathcal{S}$ space is more decoupled than the $\mathcal{W}$ space. However, the total number of channels in the style space is 9088, which increases the difficulty of detecting the channels for attribute-specific manipulation.

\begin{table}[h]
\caption{The structure of each generator layer in StyleSpace.}
\label{struct}
\begin{center}
\begin{tabular}{|c|c|c|c|}
\hline
$\mathcal{W}+$ & $\mathcal{S}$  & \# Channels with $\mathcal{S}$ & Output resolution \\
\hline
$\mathbf{w^0}$ & $\mathbf{s^0}$ & $512$ & 4x4 \\
$\mathbf{w^1}$ & $\mathbf{s^1}$ & $512$ & 4x4 \\
\hline
$\mathbf{w^1}$ & $\mathbf{s^2}$ & $512$ & 8x8 \\
$\mathbf{w^2}$ & $\mathbf{s^3}$ & $512$ & 8x8 \\
$\mathbf{w^3}$ & $\mathbf{s^4}$ & $512$ & 8x8 \\
\hline
$\mathbf{w^3}$ & $\mathbf{s^5}$ & $512$ & 16x16 \\
$\mathbf{w^4}$ & $\mathbf{s^6}$ & $512$ & 16x16 \\
$\mathbf{w^5}$ & $\mathbf{s^7}$ & $512$ & 16x16 \\
\hline
$\mathbf{w^5}$ & $\mathbf{s^8}$ & $512$ & 32x32 \\
$\mathbf{w^6}$ & $\mathbf{s^9}$ & $512$ & 32x32 \\
$\mathbf{w^7}$ & $\mathbf{s^{10}}$ & $512$ & 32x32 \\
\hline
$\mathbf{w^7}$ & $\mathbf{s^{11}}$ & $512$ & 64x64 \\
$\mathbf{w^8}$ & $\mathbf{s^{12}}$ & $512$ & 64x64 \\
$\mathbf{w^9}$ & $\mathbf{s^{13}}$ & $512$ & 64x64 \\
\hline
$\mathbf{w^9}$ & $\mathbf{s^{14}}$ & $512$ & 128x128 \\
$\mathbf{w^{10}}$ & $\mathbf{s^{15}}$ & $256$ & 128x128 \\
$\mathbf{w^{11}}$ & $\mathbf{s^{16}}$ & $256$ & 128x128 \\
\hline
$\mathbf{w^{11}}$ & $\mathbf{s^{17}}$ & $256$ & 256x256 \\
$\mathbf{w^{12}}$ & $\mathbf{s^{18}}$ & $128$ & 256x256 \\
$\mathbf{w^{13}}$ & $\mathbf{s^{19}}$ & $128$ & 256x256 \\
\hline
$\mathbf{w^{13}}$ & $\mathbf{s^{20}}$ & $128$ & 512x512 \\
$\mathbf{w^{14}}$ & $\mathbf{s^{21}}$ & $64$ & 512x512 \\
$\mathbf{w^{15}}$ & $\mathbf{s^{22}}$ & $64$ & 512x512 \\
\hline
$\mathbf{w^{15}}$ & $\mathbf{s^{23}}$ & $64$ & 1024x1024 \\
$\mathbf{w^{16}}$ & $\mathbf{s^{24}}$ & $32$ & 1024x1024 \\
$\mathbf{w^{17}}$ & $\mathbf{s^{25}}$ & $32$ & 1024x1024 \\
\hline
\end{tabular}
\end{center}
\end{table}

\section{Hierarchical representation of attributes in the $\mathcal{S}$ space}
\label{hier}
Inspired by style mixing, we demonstrate the generated images with only the front $k$ layers activation with respect to $\mathbf{s}$ codes. 	The style codes of the remaining layers come from the average codes learned by the network. As shown in Fig.~\ref{layers_activate},
the bottom layers tend to control the position and rotation of the face. The middle layers are inclined towards the local objects of the face. The top layers correspond to the color. The style code $\mathbf{s}$ modulates the average face to generate various images layer by layer. This method presents a rough explanation of the style code for each layer. 


\section{Top-$k$ channels with specific attributes}
\label{gradients_topk}

\begin{landscape}
\begin{figure}
\includegraphics[scale=0.20]{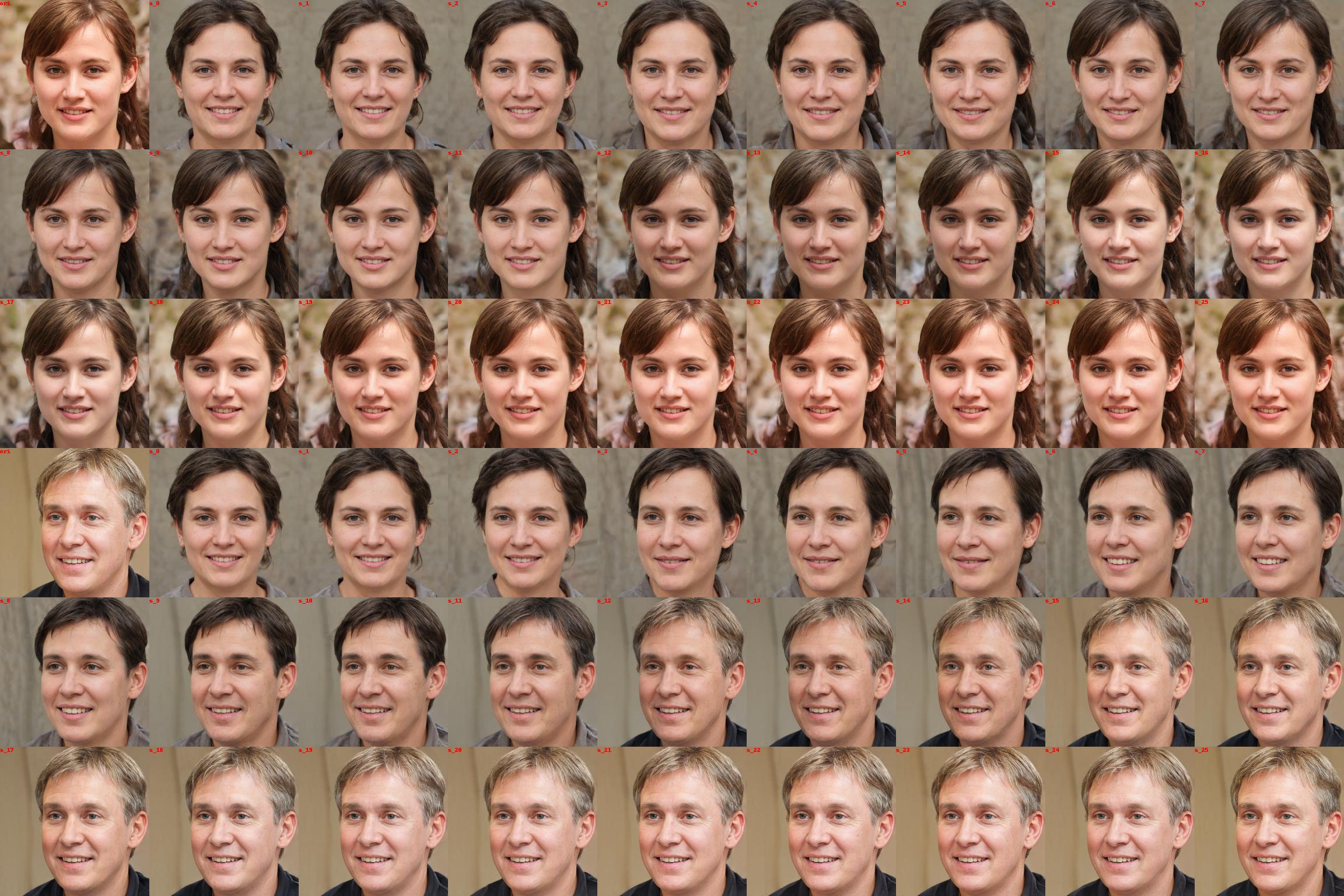}
\caption{Hierarchical representation of attributes in the $\mathcal{S}$ spaces}
\label{layers_activate}
\end{figure}
\end{landscape}

\begin{figure*}
\begin{center}
\includegraphics[scale=0.20]{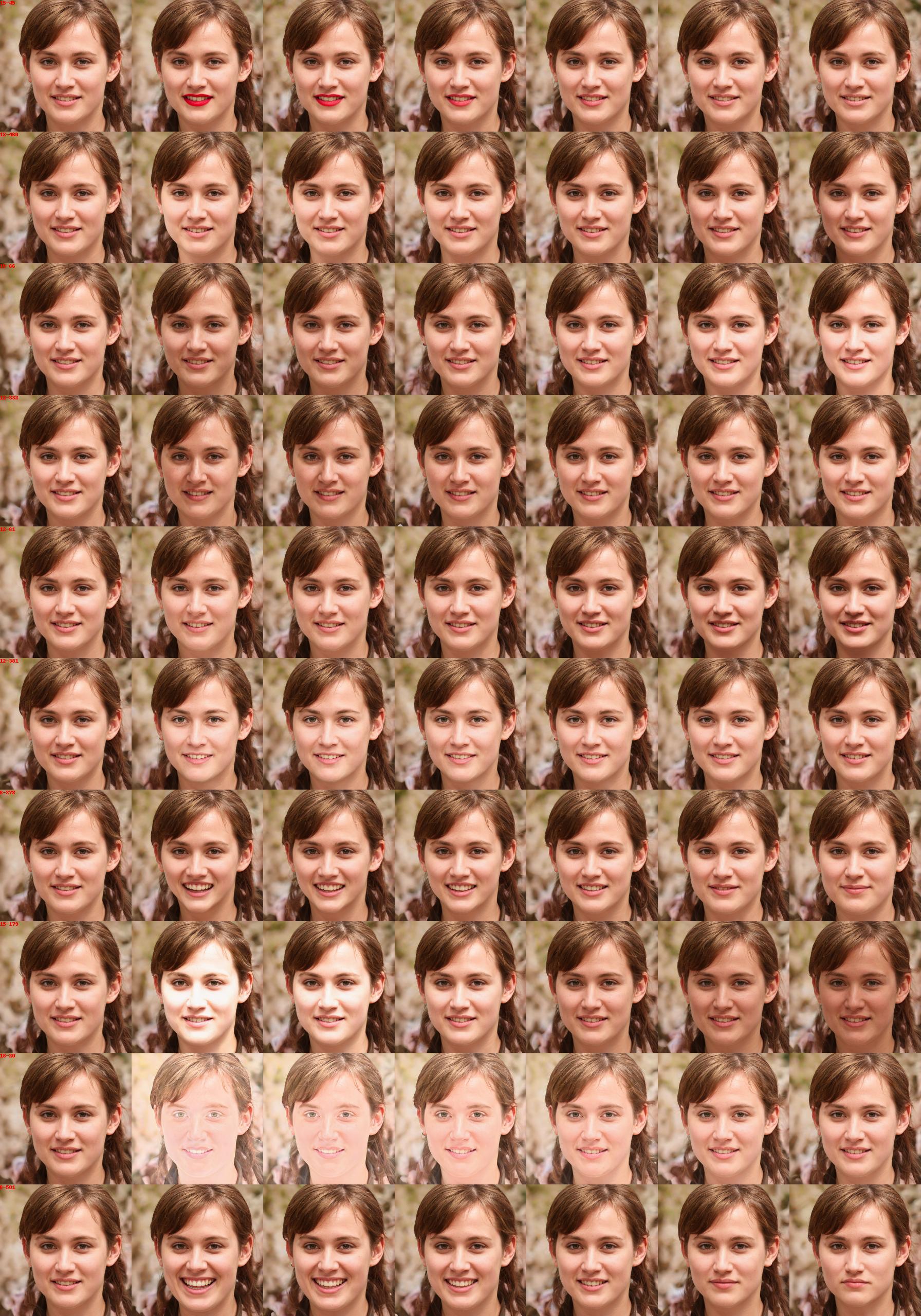}
\caption{Manipulation on the top $10$ channels of gradient intensity about mouth attribute region.  In each row, the leftmost image is the original image.}
\label{u_lip_activate}
\end{center}
\end{figure*}

\begin{landscape}
\begin{figure}
\begin{center}
\includegraphics[scale=0.23]{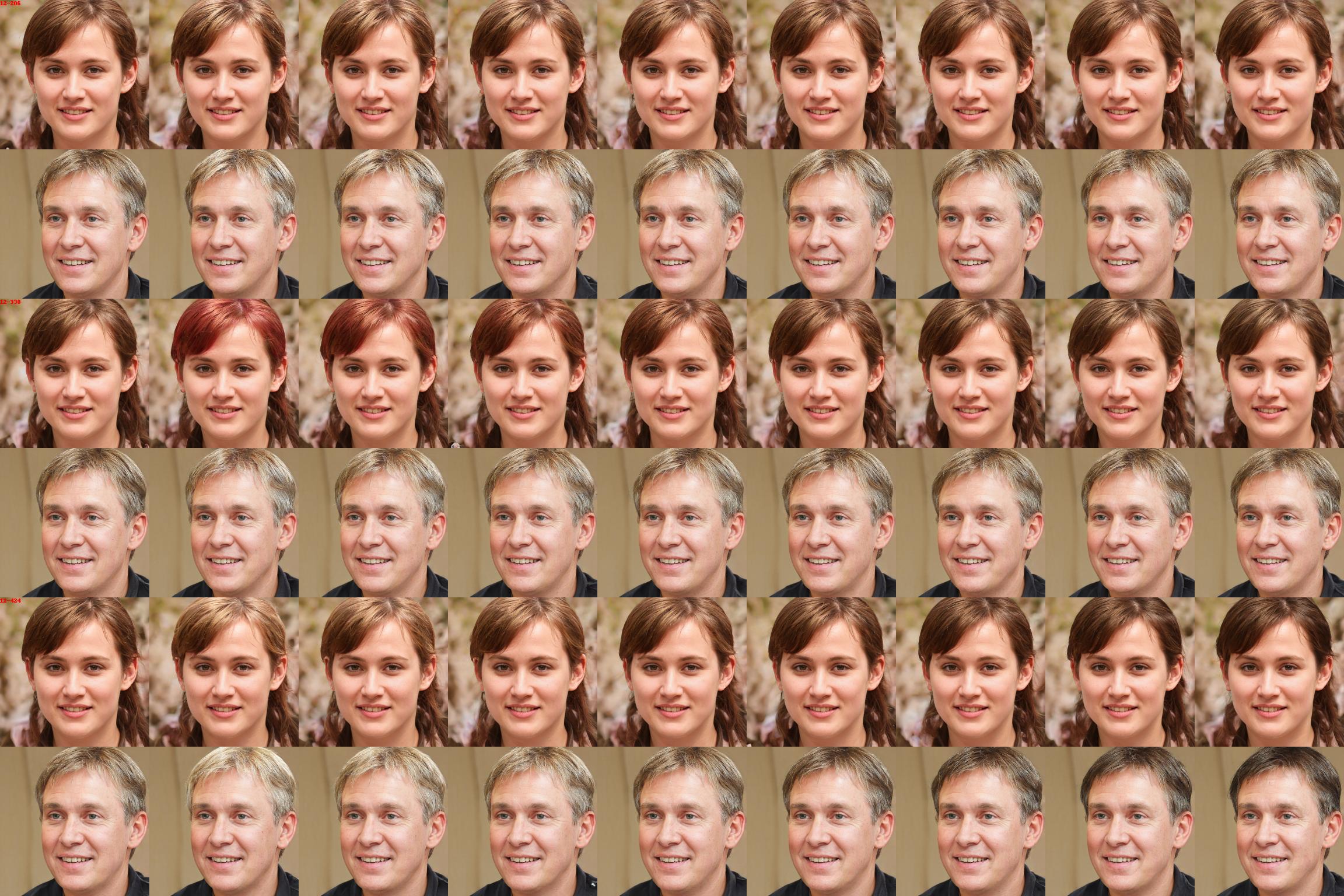}
\caption{Detection of the various channels with the mean gradient intensity about hair attribute region.}
\label{hair_ayers_activate}
\end{center}
\end{figure}
\end{landscape}

\begin{landscape}
\begin{figure}
\begin{center}
\includegraphics[scale=0.23]{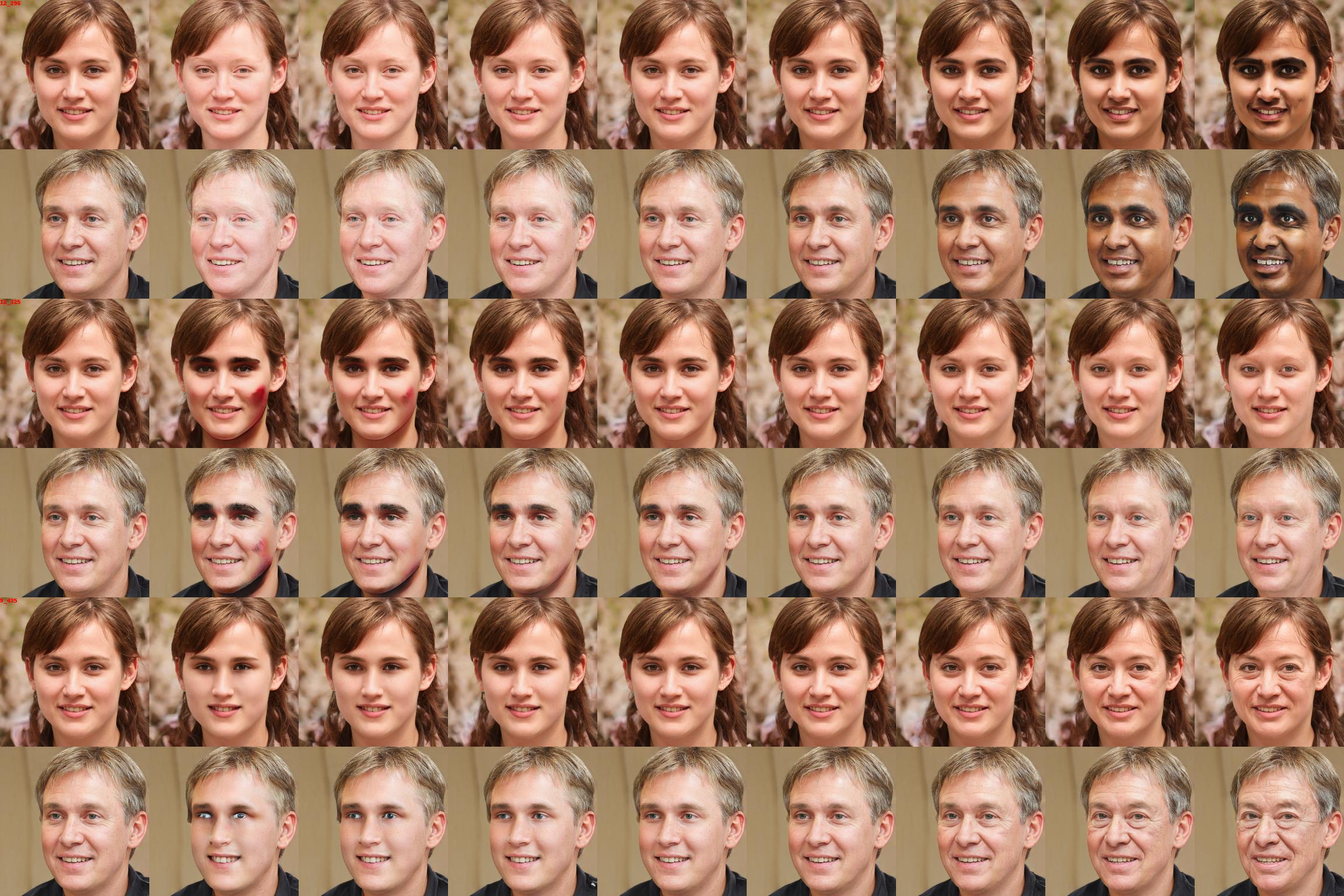}
\caption{Manipulation on the top $3$ channels from the $mean(g^a_r)$ about semantic attribute 'young'.}
\label{young_activate}
\end{center}
\end{figure}
\end{landscape}

\begin{figure*}
\begin{center}
\includegraphics[scale=0.20]{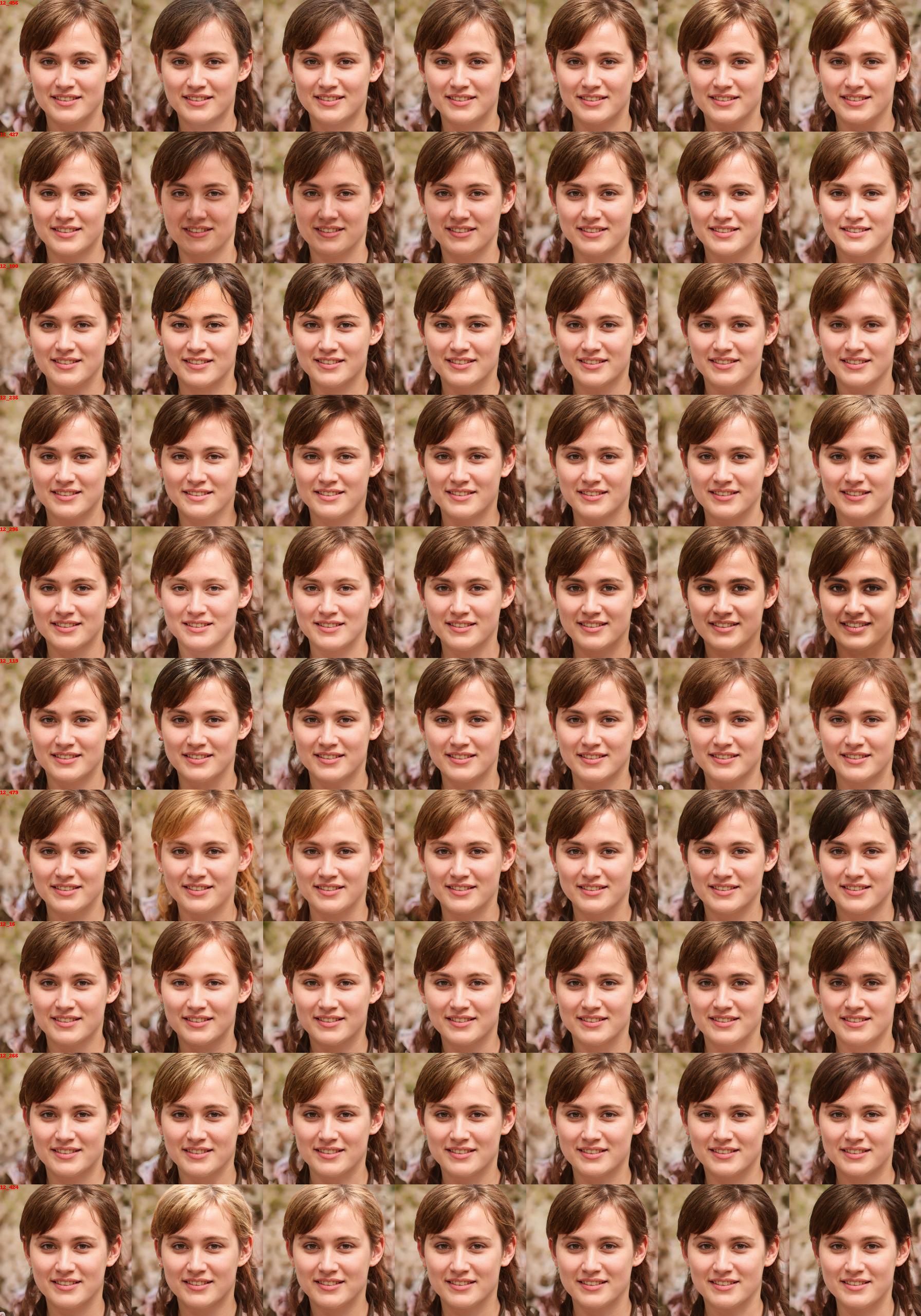}
\caption{Manipulation on the top $10$ channels of gradient intensity on $12$-th layer about semantic attribute ``Black hair''.  In each row, the leftmost image is the original image.}
\label{black_12}
\end{center}
\end{figure*}

\begin{figure*}
\begin{center}
\includegraphics[scale=0.20]{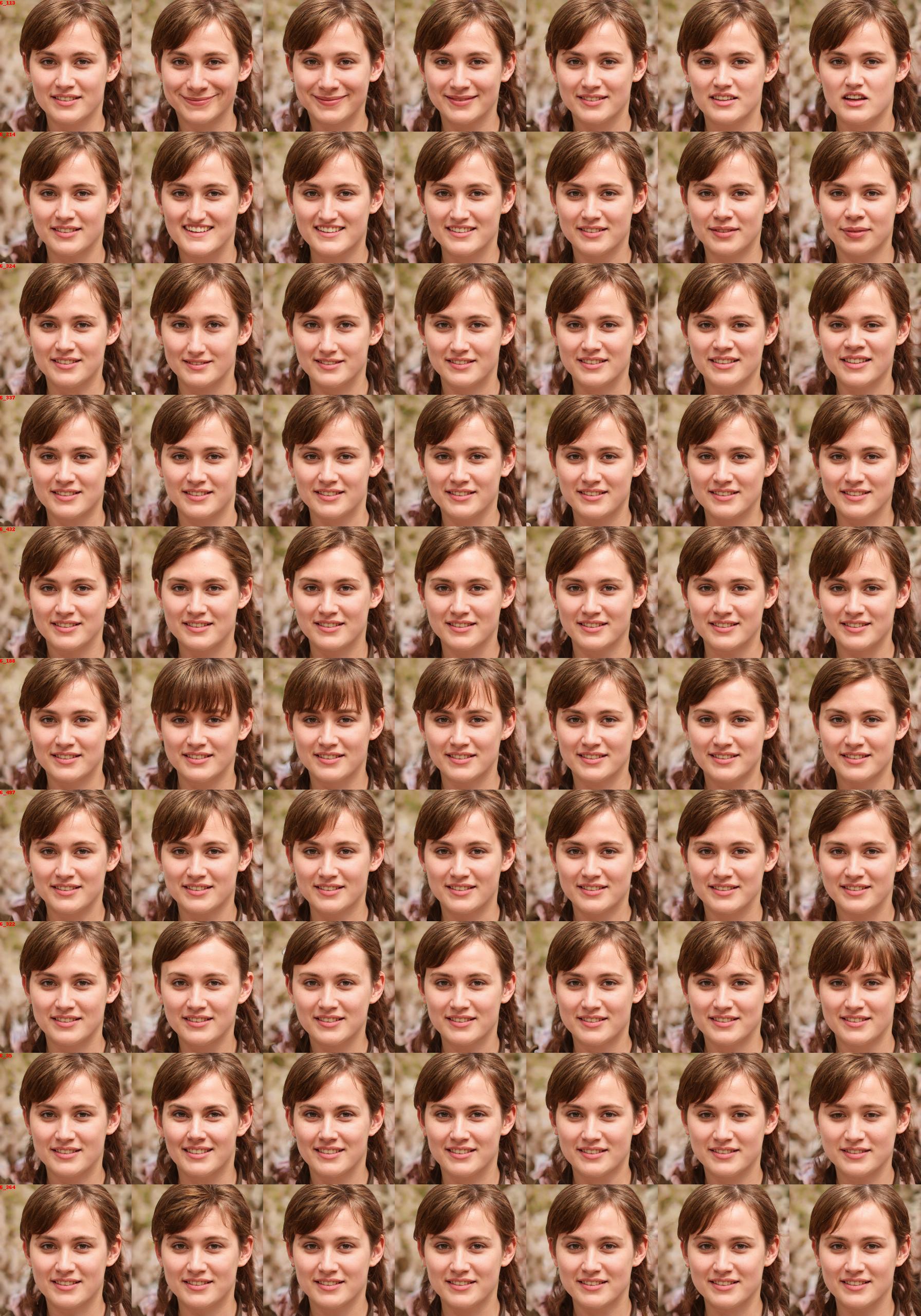}
\caption{Manipulation on the top $10$ channels of gradient intensity on $6$-th layer about semantic attribute ``Black hair''.}
\label{black_6}
\end{center}
\end{figure*}

Fig.~\ref{layers_activate} shows the edited image for the top-$10$ channels of  gradient intensities about attribute regions.  The detection channels are ordered from the bottom to the top of each figure. We make single-channel manipulation with various editing intensities in each row.
We observe the different style channels that cooperatively control the generation of the attribute region. 
For example in Fig.~\ref{u_lip_activate}, the $6$-th style layer controls the shape of the mouth, and the $10$-th and $12$-th layers are prone to control the color of the mouth.
Compared with Fig.~\ref{layers_activate} in \ref{hier}, we choose the top-$k$ channels based on the gradient intensities of each layer, which can more accurately locate layers and channels corresponding to specific attributes.
As shown in Fig.~\ref{hair_ayers_activate} of \ref{hier}, a series of style channels control the various hair attributes when detecting on the samples.
 Fig.~\ref{young_activate} shows the edited image for the top-$10$ channels of absolute values for gradients about ``young'' attribute. 
Figures \ref{black_12} and \ref{black_6} show the manipulation results about detetion channles on the 6-th and 12-th layers. In each row, the leftmost image is the original image. The other results are derived from the continuous manipulation along the editing direction.
Compare to the ``Black hair'' attribute on the figures \ref{black_12} and \ref{black_6}, the detection channels of layers are respond to the different semantic properties.


\end{document}